\documentclass[10pt]{article} %

\usepackage[preprint]{rlc}

\usepackage{amssymb}            %
\usepackage{mathtools}          %
\usepackage{mathrsfs}           %
\usepackage{graphicx}           %
\usepackage{subcaption}         %
\usepackage{url}                %
\usepackage{enumitem}

\usepackage[frozencache,cachedir=.]{minted}
\usepackage{mdframed} %
\title{%
OCALM:
Object-Centric
Assessment
\\
with Language Models%
}

\author{\\\name Timo Kaufmann\thanks{Equal contribution}$^{\ ,1,2}$, \ \textbf{Jannis Blüml}\(^{*,3,4}\), \ \textbf{Antonia Wüst}$^{*,3}$, \  \textbf{Quentin Delfosse}$^{*,3,5}$, \\  \textbf{Kristian Kersting}$^{3,4,6,7}$ \textbf{\&} \ \textbf{Eyke Hüllermeier}$^{1,2}$ \\
\addr \href{mailto:timo.kaufmann@ifi.lmu.de}{timo.kaufmann@ifi.lmu.de}, \href{mailto:quentin.delfosse@cs.tu-darmstadt.de}{quentin.delfosse@cs.tu-darmstadt.de}, \href{mailto:blueml@cs.tu-darmstadt.de}{blueml@cs.tu-darmstadt.de} \\ \\
$^1$LMU Munich, Germany \\
$^2$Munich Center of Machine Learning (MCML) \\
$^3$AI and ML Group, Technical University of Darmstadt, Germany \\
$^4$Hessian Center for Artificial Intelligence (hessian.AI) \\
$^5$National Research Center for Applied Cybersecurity (Athene) \\
$^6$Centre for Cognitive Science of Darmstadt \\
$^7$German Research Center for Artificial Intelligence (DFKI)  
}

\newtoggle{final}
\togglefalse{final}
\toggletrue{final} %
\usepackage{url}

\usepackage{booktabs}
\usepackage[utf8]{inputenc}
\usepackage{enumitem}

\usepackage{xspace}
\newcommand{\eg}{\emph{e.g.}\xspace} 

\newcommand{\ie}{\emph{i.e.}\xspace} 

\newcommand{\cf}{\emph{cf.}~}

\iftoggle{final}{%
	\usepackage[disable]{todonotes}
}{%
	\setlength{\marginparwidth}{2.5cm}
	\usepackage[textsize=tiny]{todonotes}
}

\usepackage[capitalise,noabbrev]{cleveref}

\usepackage{tabularx}
\usepackage{doi}

\begin{document}

\maketitle

\begin{abstract}
    Properly defining a reward signal to efficiently train a reinforcement learning (RL) agent is a challenging task. 
    Designing balanced objective functions from which a desired behavior can emerge requires expert knowledge, especially for complex environments.
    Learning rewards from human feedback or using large language models (LLMs) to directly provide rewards are promising alternatives, allowing non-experts to specify goals for the agent.
    However, black-box reward models make it difficult to debug the reward. 
    In this work, we propose Object-Centric Assessment with Language Models (OCALM) to derive inherently interpretable reward functions for RL agents from natural language task descriptions.
    OCALM uses the extensive world-knowledge of LLMs while leveraging the object-centric nature common to many environments to derive reward functions focused on relational concepts, providing RL agents with the ability to derive policies from task descriptions.
\end{abstract}

\section{Introduction}

Defining reward functions for reinforcement learning (RL) agents is a notoriously challenging task \citep{amodei2016concrete,knox2023reward, delfosse2024interpretable, kohler2024interpretable}.
Consequently, reward functions are often unavailable or sub-optimal, suffering from issues such as reward sparsity \citep{andrychowicz2017hindsight} or difficult credit assignment \citep{raposo2021synthetic,wu2023read}.
While standard RL benchmark are equipped with predefined reward functions, real-world tasks typically lack explicit reward signals.
Existing approaches, such as reinforcement learning from human feedback (RLHF) \citep{christiano2017deep, ouyang2022training,kaufmann2023survey}, circumvent the reward specification problem by learning a reward model from human feedback.
However, it generally requires learning reward models from scratch, which can be slow and inefficient.
Further, their \emph{black box} nature complicates the understanding and adjustment of the signal.

\begin{figure}
    \centering
    \includegraphics[width=0.85\linewidth]{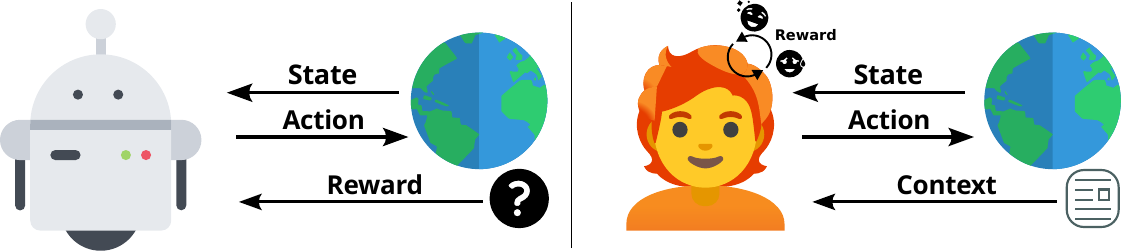}%
    \caption{%
    \textbf{Contrary to RL agents, humans infer objectives from context.}
    The RL setting assumes the existence of an external reward function, wheres humans are able to infer rewards from information about the environment and task context.
    }%
    \label{fig:motivation}%
\end{figure}

In contrast to RL agents, humans can learn to solve tasks without clear external rewards, deriving their own objectives from task context \citep{deci2013intrinsic} (\cf~\cref{fig:motivation}).
Given such context, humans formulate their own goals and generate a corresponding reward signal autonomously~\citep{spence1947role,oudeyer2008how}.
This capability stems from our rich understanding of the world, enabling us to derive specific goals from potentially vague task descriptions.
Conversely, RL agents typically lack common sense and are trained tabula rasa, devoid of any world knowledge.
In this paper, we demonstrate that large language models (LLMs) are capable of a similar feat, using their acquired world knowledge to derive goals from task descriptions that can be used by RL agents.

While previous works have demonstrated that LLMs can provide RL agents with a reward signal derived from context \citep{ma2024eureka,xie2024text2reward}, these approaches do not capitalize on the object-centric and relational nature prevalent in environments that incorporate relational reasoning challenges.
Assuming object-centricity offers a powerful inductive bias, enabling agents to reason about the world in terms of objects and their interactions rather than through raw pixels or other low-level features~\citep {delfosse2023interpretable,luo2024insight}.
We demonstrate that by directing the LLM to concentrate on the relationships between objects we can significantly enhance the effectiveness of the generated reward functions and, consequently, improve the final agent's performance.

\begin{figure}
\centering
    \includegraphics[width=0.85\textwidth]{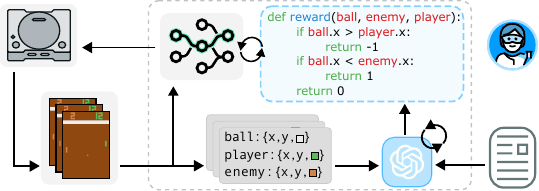}
    \caption{%
    \textbf{Object-Centric Assessment with Language Models.}
    OCALM extracts a \emph{neurosymbolic abstraction} from the raw state, provided to a language model together with the game's context, to generate a \emph{symbolic reward function} (in python).
    The language model first generates relational utility functions, that are then used in the reward function.
    This transparent reward can be inspected and used to train the policy.
    }\label{fig:method}
    \vspace{-4mm}
\end{figure}

We introduce Object-Centric Assessment with Language Models (OCALM, \cf \cref{fig:method}) as an approach to derive inherently interpretable reward functions for RL agents from the natural-language context of tasks.
OCALM leverages both the extensive world-knowledge of LLMs and the object-centric nature of many environments to equip RL agents with a rich understanding of the world and the ability to derive goals from task descriptions.
We leverage the powerful inductive bias of object-centric reasoning, directing the LLM to focus on the relationships between objects in the environment using a multi-turn interaction.
OCALM comprises two main components:
(1) a language model that generates a symbolic reward function from text-based task context, and
(2) an RL agent that trains based on this derived reward function.

In our evaluations on the iconic Atari Learning Environment (ALE)~\citep{mnih2013playing}, we provide experimental evidence of OCALM's performance, particularly its capability to learn policies comparable to those of agents trained with ground-truth reward functions.
We demonstrate the benefits of object-centric reasoning through the relational inductive bias, which significantly enhances the quality of the learned reward functions.
Additionally, we highlight the interpretability of the learned reward functions and OCALM's applicability to environments lacking ground-truth rewards.

In summary, our specific contributions are:
\begin{description}[itemsep=1pt,parsep=2pt,topsep=0pt,partopsep=2pt]
    \item[(i)]
    We introduce OCALM, an approach for inferring relational (\emph{object-centric}) reward functions from text-based task descriptions for RL agents.
    \item[(ii)]
    We show that OCALM produces learnable reward functions, that lead to RL agents performing on par with agents trained on the original reward.
    \item[(iii)]
    We empirically demonstrate the importance of object-centric reasoning for enhancing the performance of OCALM.
    \item[(iv)]
    We establish that OCALM provides inherently interpretable reward functions.
\end{description}

In the remainder of the paper, we provide a detailed description of OCALM and its components (\cref{sec:method}), followed by experimental evaluations and analysis (\cref{sec:experiments}).
We address related work (\cref{sec:related_work}) before concluding (\cref{sec:conclusion}).

\section{Object-Centric Assessment with Language Models} \label{sec:method}

OCALM provides RL agents with inherently interpretable reward functions derived from text-based task descriptions.
We follow a multistep approach, as depicted in \cref{fig:method}, to achieve this goal.
\begin{description}[itemsep=4pt,parsep=4pt,topsep=0pt,partopsep=2pt,leftmargin=1pt]
    \item[(1) Context definition.]
        We start by gathering a \emph{natural-language task description} and extracting an \emph{object-centric state abstraction} from the raw input state.
        The task descriptions (listed in \cref{app:llm_results}) are based on the short descriptions of each Atari environment~ \citep{towers2023gymnasium}, slightly modified to add missing information.
        The object-centric state abstractions include the properties of each object, such as their class, position, size, and color.
        It is given by the classes provided by the OCAtari framework \citep{delfosse2023ocatari}, \ie, the parent game object class and the game-specific objects (examples are provided in \cref{app:game-objects}).
        Game objects related to the score were omitted, since we assume a reward-free environment.
        The task description and the object-centric state abstraction form the task context, which is provided to the language model.
    \item[(2) LLM-driven reward generation.]
        The large language model (LLM) processes the task context to generate a symbolic reward function in the form of Python source code.
        We use a guided multi-turn approach to direct the LLM to focus on the relationships between objects in the environment, similar to chain-of-thought reasoning~\citep{wei2022chainofthought}.
        \begin{description}[itemsep=0.2mm]
            \item[(2.1) Relational concept extraction.]
                The LLM is tasked with generating relational functions that describe the relationships between objects in the environment (\cf \cref{lst:cot_prompt_functions}, \cref{app:llm_details}), which are important to understand the game states.
            \item[(2.2) Reward generation.]
                Given the task context and the created utility functions, the LLM generates a symbolic reward function (\cf \cref{lst:cot_prompt_reward}, \cref{app:llm_details}).
            \item[(2.3) Reward scaling.]
                As a last step, we prompt the LLM (\cref{lst:reward_scaling}, \cref{app:llm_details}) to adjust the created reward function in such a way that the rewards are on a scale from $-1$ to $1$.
        \end{description}
        The resulting reward function is a Python function mapping the object-centric state abstraction to a scalar reward with semantic descriptions. This code is interpretable, allowing experts to inspect and verify it before proceeding. We also present an ablated version, OCALM (no relations), where the LLM generates the reward function directly, skipping the relational and reward scaling steps.
        While the no relations version may still use relational concepts, we do not prompt it to do so.
        We use a modified prompt (\cref{lst:direct_prompt} in \cref{app:llm_details}) in that case.
        We provide a shortened example of a generated reward function in \cref{lst:cot_freeway} and the full reward functions in \cref{app:llm_results}.
    \item[(3) Policy training.]
        The derived reward function is used to train an RL agent, which learns a policy that maximizes the reward.
        The agent can be trained using any conventional RL algorithm, Proximal Policy Optimization \citep[PPO,][]{schulman2017proximal} in our experiments.
\end{description}

\begin{listing}[t!]
\caption{An example reward function generated by OCALM (full). Implementation of relational utility function elided and unused utilities removed. The full version is in \cref{lst:cot_freeway}.}
\vspace{-3mm}
\label{lst:rel_freeway_trimmed}
\begin{minted}{python}
def detect_collision(chicken, car): [...]
def has_reached_top(chicken, screen_height): [...]
def progress_made(chicken, screen_height): [...]

def reward_function(game_objects) -> float:
    # Initialize reward
    reward = 0.0

    # Constants
    SCREEN_HEIGHT = 160
    COLLISION_PENALTY = -1.0  # Scaled down to fit within [-1, 1]
    PROGRESS_REWARD = 0.1  # Scaled down to incrementally increase reward
    SUCCESS_REWARD = 1.0  # Maximum reward for reaching the top

    # Filter out chickens and cars from game_objects
    chickens = [obj for obj in game_objects if isinstance(obj, Chicken)]
    cars = [obj for obj in game_objects if isinstance(obj, Car)]

    # Assume control of the leftmost chicken (player's chicken)
    if chickens:
        player_chicken = min(chickens, key=lambda c: c.x)

        # Check if the chicken has reached the top
        if has_reached_top(player_chicken, SCREEN_HEIGHT):
            reward += SUCCESS_REWARD

        # Reward based on progress towards the top
        reward += progress_made(player_chicken, SCREEN_HEIGHT) * PROGRESS_REWARD

        # Check for collisions with any car
        for car in cars:
            if detect_collision(player_chicken, car):
                reward += COLLISION_PENALTY
                break  # Only penalize once per time step

    # Ensure reward stays within the range [-1, 1]
    reward = max(min(reward, 1.0), -1.0)

    return reward
\end{minted}
\end{listing}

\section{Experimental Evaluation}\label{sec:experiments}

\textbf{Experimental setup:}
We evaluate OCALM on four Atari games (Pong, Freeway, Skiing, and Seaquest) from the ALE \citep{bellemare2013arcade}. All results are averaged over three seeds for each agent configuration, with standard deviation indicated.
We use Proximal Policy Optimization \citep[PPO,][]{schulman2017proximal} as the base architecture due to its success in Atari games.
The input representation is a stack of four gray-scaled $84\times84$ images, introduced by \cite{mnih2015humanlevel}.
All agents are trained using 10M frames with the implementation by \citet{huang2022cleanrl} and default hyperparameters (\cf \cref{app:hyperparameters}).
We compare our OCALM agents trained with the 'true' reward functions given by the ALE environment, typically based on game score.
Both types of agents are evaluated against the true game score.
All evaluations use the latest \textit{v5} version of the ALE environments, following best-practices to prevent overfitting \citep{machado2018revisiting}.
The results are presented as figures, refer to \cref{app:num_results} for numerical results. To generate our reward function, we assume access to object-centric state descriptions of the game state.
To focus on description-based reward derivation, we use representations from the Object-Centric Atari (OCAtari) framework~\citep{delfosse2023ocatari}.
While a learned object detector could extract objects from raw input~\citep{redmon2016you,lin2020space}, we use OCAtari for simplicity.  %

We evaluate the OCALM approach to answer the following research questions:
\begin{description}[itemsep=0pt,parsep=2pt,topsep=0pt,partopsep=0pt]
    \item[\textbf{(Q1)}]
    Does OCALM generate rewards that correspond to learnable tasks?
    \item[\textbf{(Q2)}]
    Can OCALM agents master Atari environments without access to the true game score?
    \item[\textbf{(Q3)}]
    How does relation-focused reward derivation influence performance and interpretability?
    \item[\textbf{(Q4)}]
    How interpretable are the reward functions generated by OCALM agents?
\end{description}

\begin{figure}
    \centering
    \begin{subfigure}[b]{\textwidth}
        \includegraphics[width=\linewidth]{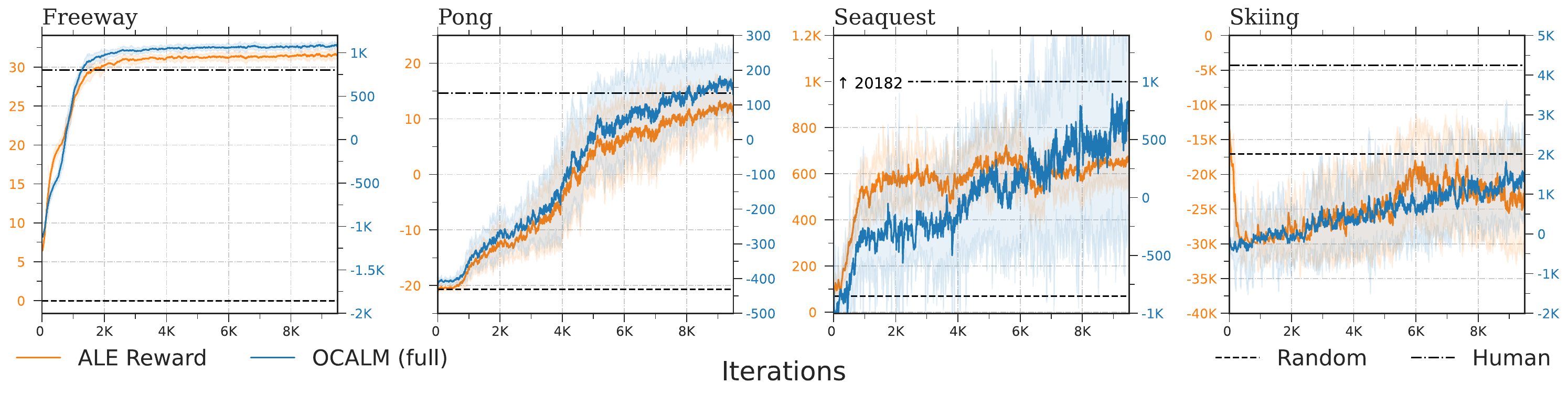}
        \caption{Contrasting OCALM (full) rewards with environment rewards.}\label{fig:ocalm_full_correlation}
    \end{subfigure}
    \\
    \vspace{5mm}
    \begin{subfigure}[b]{\textwidth}
        \centering
        \includegraphics[width=\linewidth]{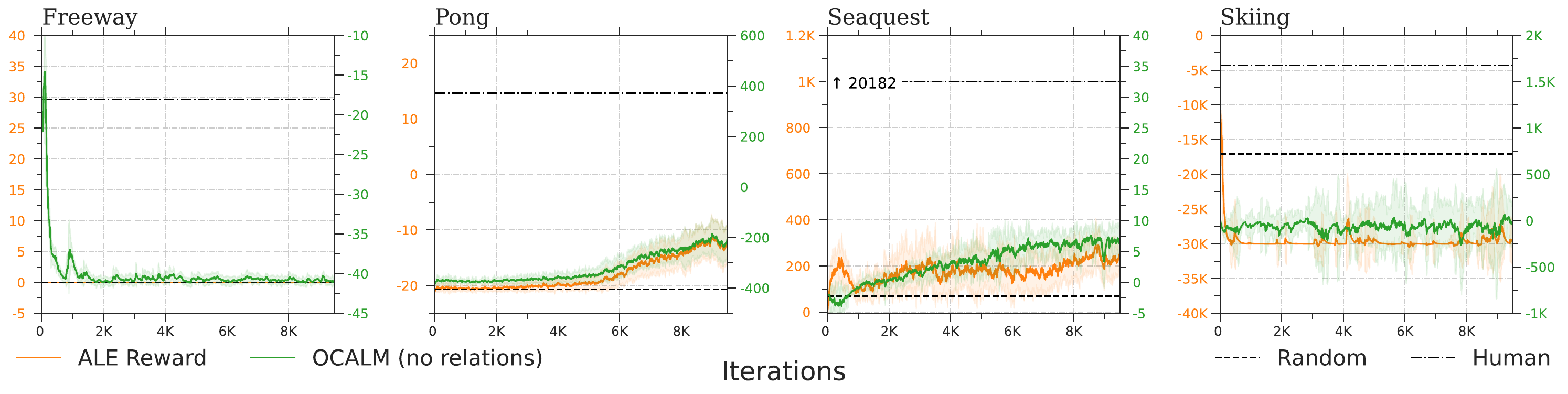}
        \caption{Contrasting OCALM rewards (no relations) with environment rewards.}\label{fig:ocalm_no_relations_correlation}
    \end{subfigure}
    \vspace{0mm}
    \caption{%
        \textbf{%
        OCALM generates meaningful reward functions that correlate with the intended game rewards.
        }
        These figures show the performance of agents trained on OCALM-derived rewards, measured on both the OCALM-derived reward and the environment reward.
        The scales of rewards differ, therefore the axes are scaled to better visualize the correlation.
        Both plot for the same game share the same axis range for better comparability.
        The results indicate that (1) the reward functions generated by OCALM correspond to objectives learnable by an RL agent, and (2) the OCALM-derived rewards correlate with the environment rewards.
        All experiments were averaged over $3$ seeds, with standard deviations shown as shaded areas.
    }\label{fig:results_ocalm}
    \vspace{-4mm}
\end{figure} 

\textbf{OCALM generates reward signals allowing to master the game (Q1).}
We first test whether OCALM generates rewards that correspond to learnable tasks.
For this purpose, we track the learning curves of agents trained on OCALM-derived rewards and verify that agents improve over time, e.g., learn to maximize the reward.
\Cref{fig:results_ocalm} shows that this is generally the case, with an exception for Freeway when using the ablated variant of OCALM (no relations) (see \cref{fig:ocalm_no_relations_correlation}).
For all other games, and for all games when using the full OCALM pipeline, the agents improve over time when measured on the OCALM-derived reward.
Without the relational inductive bias, OCALM fails to generate learnable rewards for Freeway.
This is due to a bug in the generated reward function (see \cref{app:freeway-rewards}, \cref{lst:direct_freeway}), which fails to identify the player-controlled chicken.
Although it is quite possible that the relational inductive bias helps to avoid such bugs through mechanisms similar to chain-of-thought reasoning, they cannot entirely be prevented.
More research is necessary to understand the impact of the relational inductive bias on the failure rate of generated reward functions.
Iterative refinement could help further alleviate this issue, but generating successful reward functions in a single shot remains a significant computational advantage.

\begin{figure}
    \centering
    \includegraphics[width=\linewidth]{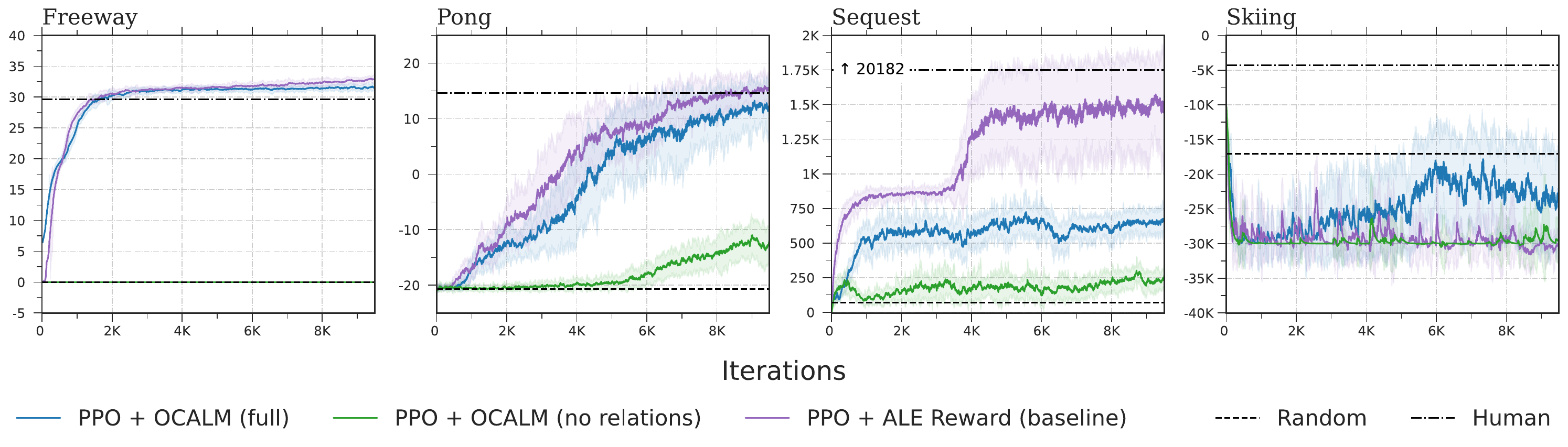}
    \caption{%
    \textbf{OCALM agents can master different Atari environments.}
    Comparing the performance of agents trained on OCALM-derived rewards to agents trained on the true game score.
    All experiments were averaged over $3$ seeds, with standard deviations shown as shaded areas.
    }\label{fig:results_ale}
\end{figure} 
\textbf{OCALM-based agents can master different Atari environments without access to the true game score (Q2).}
\Cref{fig:results_ale} shows the performance of agents trained on OCALM-derived rewards compared to those trained on the true game score.
Performance is measured on the true game score in both cases, which OCALM agents cannot access during training.
Our goal is not to exceed the baseline agents' performance, but to show that OCALM agents can master environments without access to true rewards.
Even though the OCALM-derived reward functions differ from the environment reward, we observe that OCALM agents, when using relational prompting, reliably improve their performance over the course of training when measured on the true game score.
This further confirms that the reward functions generated by OCALM are correlated with the true game score, as discussed in the previous paragrpah.
For Freeway and Seaquest in particular, OCALM agents were able to reach competitive performance compared to the baseline agents, without requiring access to the true game score.
Although OCALM agents do not match the baseline's performance in Pong, they still show significant learning progress, again indicating the reward function generated by OCALM correlates with the environment reward.

\textbf{Relational prompting of OCALM agents improves reward quality (Q3).}
\Cref{fig:results_ale} shows that agents trained on OCALM-derived rewards with the relational inductive bias (denoted OCALM (full)) generally outperform those without it (denoted OCALM (no relations)).
This is particularly evident in Freeway and Seaquest, where OCALM (full) agents reach performance competitive with the baseline, while OCALM (no relations) agents fail to learn the task.
In Pong and Skiing, OCALM (full) agents perform equivalently to OCALM (no relations) agents, indicating the relational inductive bias is not beneficial in all cases, but also does not harm performance.
Note that the OCALM (no relations) variant also skips the reward scaling step, which could be another contributing factor to the performance difference.
Qualitatively, when inspecting the reward functions generated by OCALM (\cref{app:llm_results}), we observe that the relational inductive bias helps to capture more complex concepts, such as the distance to the nearest obstacle in Skiing \cref{app:skiing-rewards}, which in turn can lead to better-shaped reward functions.

\textbf{OCALM generates interpretable reward functions (Q4).}
The reward functions generated by OCALM (\cf \cref{app:llm_results}) are based on high-level objects and relations, documented with comments, making them easy to interpret and understand.
Relational prompting further aids in generating interpretable reward functions by introducing easy-to-understand relational concepts, which add an abstraction layer to the reward function.
Examples include collision detection, a relation generated for all games, and easily understandable concepts such as \verb|has_reached_top| in Freeway (\cref{app:freeway-rewards}), and more complex relations including multiple objects such as \verb|detect_score_event| in Pong (\cref{app:pong-rewards}) or \verb|check_gate_passage| in Skiing (\cref{app:skiing-rewards}).

\section{Limitations}
In our evaluations, we use the integrated object extractor of OCAtari which provides ground truth data.
Such extractors can also be optimized using supervised \citep{redmon2016you} or self-supervised \citep{lin2020space, delfosse2023boosting} object detection methods.
We additionally rely on the language models ability to generate a reward function in a single shot.
While our relational inductive bias helps, the LLM may miss crucial information such as the frequency of certain events, which is important to tune the relative scales of different reward components.
Related works \citep{ma2024eureka,xie2024text2reward} rely on iterative refinement of the reward function, which could further enhance OCALM's performance.
Nonetheless, the relational inductive bias enables OCALM to frequently learn successful reward functions in a single shot, a significant computational efficiency advantage.

\section{Related Work}\label{sec:related_work}

\begin{table}
    \caption{%
    Relating OCALM to the most closely related work, EUREKA~\citep{ma2024eureka} and Text2Reward~\citep{xie2024text2reward}.
    In contrast to our work, EUREKA and Text2Reward use mulitple iterations to refine the reward function.
    All three approaches provide the LLM with additional information about the environment and the task.
    While EUREKA and Text2Reward evaluate on joint control tasks (locomotion and manipulation), we evaluate on relational tasks (Atari games).
    }\label{tbl:related_work}
    \centering
    \begin{tabular}{lcccc}
        \textbf{Approach} & \textbf{1-Shot} & \textbf{Add.\ Context} & \textbf{Relational} & \textbf{Evaluation} \\
        \toprule
        EUREKA & No & Source code & No & Joint control \\
        Text2Reward & No & Symb.\ state abstr. & No & Joint control \\
        OCALM (\textbf{ours}) & Yes & Symb.\ state abstr. & Yes & Relational tasks \\
    \end{tabular}
\end{table}

OCALM lies at the intersection of several research areas, including reinforcement learning from human feedback (RLHF), language-guided RL, explainable RL and relation extraction.

\textbf{Reward learning}
has been studied in various forms and based on different sources, such as demonstrations~\citep{arora2021survey} and human preferences~\citep{kaufmann2023survey}.
While these approaches can be very effective, they often require a large amount of human supervision, which can be costly and time-consuming.
Our method, by combining human guidance given through the task description with the extensive world knowledge of LLMs, helps to alleviate this issue.
Particularly closely related to ours, RL-VLM-F~\citep{wang2024rlvlmf} is a notable approach that learns a reward model from pairwise comparisons judged by a vision-language model based on a natural language task description.
Similar to OCALM, this leverages the prior knowledge of the vision-language model.
In contrast to our work, however,~\citet{wang2024rlvlmf} and most other reward learning methods learn black-box reward models in the form of neural networks, which are not interpretable.

\textbf{LLM-written reward functions}
have been studied by~\citet{ma2024eureka}, who propose EUREKA, and~\citet{xie2024text2reward}, who propose Text2Reward.
These are the most closely related works to ours.
\Cref{tbl:related_work} highlights the most relevant differences between our method and theirs.
Like our approach, EUREKA and Text2Reward use LLMs to generate reward functions for RL agents.
They assume access to a natural language task and  environment descriptions, specifying the observations' structure.
EUREKA assumes descriptions are given in the form of incomplete source code, while Text2Reward requires class definitions that define the components of the state.
Both approaches evaluate the generated reward functions on robotic manipulation and locomotion tasks.\\
EUREKA and Text2Reward work iteratively, \ie the reward function is refined based on feedback from the environment or a human expert.
This can help further improve the reward function, but also requires more computational resources, time and supervision (either from a human expert or a success signal).
Since the focus of our study is on the benefits of object-centric reward specifications, we leave the iterative refinement for future work and instead focus on improving single-shot performance.

In contrast to these prior works, we focus on relational reasoning environments, which require the agent to reason about multiple objects and their interactions.
We leverage a relational inductive bias for improved one-shot performance, reducing the need for iterative refinement and human supervision.
We further evaluate on the prominent Atari Learning Environment~\citep{bellemare2013arcade}, the most used benchmark for reinforcement learning~\cite{delfosse2023ocatari}, and show the importance of object-centric inductive biases for learning reward functions in this setting.

\textbf{RL from natural language task descriptions}
is extensively studied~\citep{nair2021learning, zhou2021inverse, pang2023natural, karalus2024tell}.
While these approaches are similar to ours in that they use natural language to specify the task, they typically do not leverage the world knowledge of LLMs, do not learn interpretable reward functions, and do not use relational inductive biases.

\textbf{Explainable RL (XRL)} is a subfield of explainable AI (XAI)~\citep{milani2023explainable,dazeley2023explainable,krajna2022explainability}.
XRL aims to offer insights into the behavior of RL agents, aiding in realigning agents.
OCALM helps in this endeavor by providing inherently interpretable reward functions, which can be inspected and verified by experts. This can be used to align the reward functions with certain societal values, such as more pacific gameplays in \eg shooting games.

\textbf{Relation extraction}
has been studied in many forms, including prior task-knowledge integration~\citep{reid2022can}, or human intervention~\citep{steinmann2023learning, wust2024pix2code, stammer2024neural} and neural guidance from a pretrained fully deep agent~\citep{delfosse2023interpretable}, based on first order logic, notably from~\citep{shindo2021differentiable, shindo2023alpha}, relying on human predicate or automated predicate invention~\citep{sha2023neural, sha2024expil}.
In contrast to these works, we use an LLM to extract relations between objects in the environment, which are then used to derive reward functions.
Particularly relevant is the work by~\citet{wu2023read}, who extract relevant relations using LLMs with access to an instruction manual.
This differs from our work in that they use the extracted relations to supplement existing rewards instead of entirely replacing the environment reward function.

\section{Conclusion}\label{sec:conclusion}

We have presented OCALM, a novel approach for deriving inherently interpretable reward functions for RL agents from natural language task descriptions.
Our method leverages the extensive world knowledge of LLMs and the object-centric, relational nature of the environment to generate symbolic reward functions that can be inspected and verified by experts.
We have shown that OCALM can be used to train RL agents on Atari games, demonstrating that the derived reward functions are effective in guiding the agent to learn the desired behavior.
OCALM agents utilize the abstracted knowledge of LLMs alongside explicit relational concepts to derive effective and inherently interpretable reward functions for complex RL environments.

\subsubsection*{Broader Impact Statement}
\label{sec:broaderImpact}
We here develop RL agents with transparent, human-defined objectives, improving RL accessibility to non-experts.
We thus reduce the barrier to entry for non-experts, helping to ensure that the objectives of the agents are aligned with the user's intentions.
A malicious user can, however, utilize such approaches for developing agents with harmful objectives, thereby potentially leading to a negative impact on further users or society as a whole.
Even so, the inspectable nature of transparent approaches will allow to identify such potentially harmful misuses, or hidden misalignment. 

\subsubsection*{Acknowledgments}
\label{sec:ack}

This work is supported by LMUexcellent, funded by the Federal Ministry of Education and Research (BMBF) and the Free State of Bavaria under the Excellence Strategy of the Federal Government and the Länder and by the Hightech Agenda Bavaria.
Additionally, we thank the following organizations for funding this research project.
The German Federal Ministry of Education and Research, the Hessian Ministry of Higher Education, Research, Science and the Arts (HMWK) within their joint support of the National Research Center for Applied Cybersecurity ATHENE, via the ``SenPai: XReLeaS'' project as well as their cluster project within the Hessian Center for AI (hessian.AI) ``The Third Wave of Artificial Intelligence - 3AI''.
Further was it supported by the Priority Program (SPP) 2422 in the subproject “Optimization of active surface design of high-speed progressive tools using machine and deep learning algorithms“ funded by the German Research Foundation (DFG).

\bibliography{main}
\bibliographystyle{rlc}

\appendix

\section{Appendix}
As mentioned in the main body, the appendix contains additional materials and supporting information for the following aspects:
the hyperparameters used in this work (\cref{app:hyperparameters}),
details on the prompts used for the LLM (\cref{app:llm_details}) as well as the generated reward functions (\cref{app:llm_results}),
and
numerical results (\cref{app:num_results}).

\subsection{Hyperparameters and Experimental Details}\label{app:hyperparameters}

In this section, we list the hyperparameters used during the training and optimization of our models.
For our experiments, we adopted the parameter set proposed by \citet{huang2022cleanrl} for our PPO agents, as detailed in \cref{tab:hyperparams1}.

\begin{table}[H]
\caption{\textbf{Hyperparameter Configuration for Training Settings (PPO).} This table provides a comprehensive overview of the essential hyperparameters utilized in our experimental section.}\label{tab:hyperparams1}
\centering
\begin{tabular}{llll}
\toprule
\textbf{Hyperparameter} & \textbf{Value} & \textbf{Hyperparameter} & \textbf{Value}  \\
\midrule
batch size              & 1024           & Clipping Coef.          & 0.1               \\
$\gamma$                & 0.99           & KL target               & None                \\
minibatch size          & 256            & GAE $\lambda$              & 0.95               \\
seeds                   & 42,73,91       & input representation    & 4x84x84            \\
total timesteps         & 10M            & gym version             & 0.28.1            \\
learning rate           & 0.00025        & pytorch version         & 1.12.1            \\
optimizer               & Adam           &  \\
more information        & \multicolumn{3}{l}{\small \url{https://docs.cleanrl.dev/rl-algorithms/ppo/}}\\

\bottomrule
\end{tabular}
\end{table}

OCALM-based agents use the same PPO hyperparameter values as agents trained on ALE rewards.
All agents use ConvNets \citep{mnih2015humanlevel} with ReLU activation functions for policy and value networks.
We utilized a decreasing learning rate of $2.5 \times 10^{-4}$ over 10 million steps.
We use the Atari environments in version \texttt{v5} provided by Gymnasium \citet{towers2023gymnasium}, following best-practices recommended by \citet{machado2018revisiting}.
To accelerate training, we used 8 parallel game environments.

To mitigate noise and fluctuations, we use exponential moving average (EMA) smoothing in \cref{fig:results_ocalm} and \cref{fig:results_ale}.
We use an effective window size of \( 50 \), resulting in a smoothing factor \( \alpha = 2 / (1 + 50) \approx 0.039 \) used in the following formula:
\begin{align}
     \mathit{EMA}_t = (1-\alpha) \cdot \mathit{EMA}_{t-1} + \alpha \cdot y_t
     \,\text.
\end{align}
To manage irregular training intervals due to rewards are not always being reported in the same timestep, we ignore missing values when computing the average, relying on the EMA smoothing to provide a continuous curve.
For the error bands, we used the standard deviation of your data within a rolling window.

\subsection{Numerical Results}\label{app:num_results}

In this section, we provide additional numerical results for the experiments conducted in this work.

\begin{table}[H]
\centering
\caption{\
Numerical results for the experiments we conducted, including random and human baselines from \citet{hasselt2016deep} for comparison.
Standard deviations are provided where available.
Our agents use PPO and \textit{ALE v5} and have been trained using $10$ million frames.
The results reported are the in-game rewards from the ALE/emulator, not from OCALM.
Note, \citet{hasselt2016deep} predate the \textit{v5} version of the ALE environments used by us, which is based on the best practices outlined by \citet{machado2018revisiting}.
However, this should not change the values much since these changes have less influence on humans or the random agent.
}

\resizebox{1.0\textwidth}{!}{
\begin{tabular}{@{}lccc|cc@{}}
\toprule
\textbf{Game}
& \multicolumn{1}{c}{\begin{tabular}[c]{@{}c@{}}\textbf{PPO }\\ALE Reward (Baseline)\end{tabular}}
& \multicolumn{1}{c}{\begin{tabular}[c]{@{}c@{}}\textbf{PPO }\\OCALM (full) (Ours)\end{tabular}}
& \multicolumn{1}{c}{\begin{tabular}[c]{@{}c@{}}\textbf{PPO }\\OCALM (no relations) \end{tabular}}
& \multicolumn{1}{c}{\begin{tabular}[c]{@{}c@{}}\textbf{Random}\\van Hasselt et al.\end{tabular}}
& \multicolumn{1}{c}{\begin{tabular}[c]{@{}c@{}}\textbf{Human}\\van Hasselt et al.\end{tabular}}
\\
\midrule

\textbf{Freeway}      & $33.8\pm0.2$    &  $32.35\pm0.25$    & $0.00$  &  $0.00$    &  $29.6$\\
\textbf{Pong}         & $17.5\pm0.5$    &  $16.4\pm1.4$    & $-15.8\pm3.5$  &   $-20.7$   & $14.6$  \\
\textbf{Seaquest}     & $1132.4\pm271.4$ & $672.2\pm28.3$     & $243\pm86.3$   & $68.4$    & $20182$ \\
\textbf{Skiing}       & $-23921.3\pm10528.6$ & $-28577.7\pm2842.3$      & $-30000$  &  $-17098$  & $-4336$  \\

\bottomrule
\end{tabular}}
\label{tab:robustness1}
\end{table}

\subsection{LLM Prompting Details}\label{app:llm_details}

In our experiments we used the LLM gpt-4-turbo\footnote{\url{https://platform.openai.com/docs/models/gpt-4-turbo-and-gpt-4}} with seed $42$ and $\mathrm{top\_k} = 0$.
We further defined a system prompt that was used for both, direct and relational multi-turn prompting (\cref{lst:system_prompt}).
\begin{listing}[h]
\caption{System prompt provided to the LLM.}
\label{lst:system_prompt}
\begin{minted}[mathescape,
               linenos,
               numbersep=5pt,
               gobble=0,
               frame=lines,
               framesep=2mm,
               fontsize=\footnotesize,
               breaksymbolleft={},
               tabsize=2,breaklines]{text}
You are a helpful assistant that creates reward functions for reinforcement learning researchers.
\end{minted}
\end{listing}

For direct prompting, we asked the LLM to create a reward function directly, given a game instruction and the game object classes.
The game instructions are a few sentences that describe the objective of the game (based on the documentation of the Gymnasium environment collection \citep{towers2023gymnasium}) and the game objects are Python classes provided by the OCAtari framework \citep{delfosse2023ocatari}, further described in \cref{app:game-objects}.

\begin{listing}[h]
\caption{Prompt for reward function based on game instructions and game objects provided to the LLM.}
\label{lst:direct_prompt}
\begin{minted}[mathescape,
               linenos,
               numbersep=5pt,
               gobble=0,
               frame=lines,
               framesep=2mm,
               fontsize=\footnotesize,
               breaksymbolleft={},
               tabsize=2,breaklines]{text}
We want to create a object centric reward function to train a reinforcement learning agent to play the game <GAME>. Here is a description of the game and its objects:

<PARENT GAME OBJECT CLASS> 

<GAME OBJECT CLASSES> 

The game instructions are the following: 
 <INSTRUCTIONS> 

Please provide a Python file with a reward function that uses a list of objects of type GameObject as input that will help the agent to play the game, i.e.:
```python
def reward_function(game_objects) -> float:
    ... 
    return reward
```

Do not use undefined variables or functions. Do not give any textual explanations, just generate the python code. If you give an explanation, please provide it in the form of a comment in the code.
\end{minted}
\end{listing}

For relational multi-turn prompting, we first ask the LLM to provide functions that might be relevant for understanding the state and events of the game \cref{lst:cot_prompt_functions}. Based on these functions, the model is then asked to create a reward function with \cref{lst:cot_prompt_reward}. As a last step, the model is asked to adjust the rewards so that they are on a scale from $-1$ to $1$ \cref{lst:reward_scaling}.

\begin{listing}[h]
\caption{Prompt for helpful functions based on game instructions and game objects.}
\label{lst:cot_prompt_functions}
\begin{minted}[mathescape,
               linenos,
               numbersep=5pt,
               gobble=0,
               frame=lines,
               framesep=2mm,
               fontsize=\footnotesize,
               breaksymbolleft={},
               tabsize=2,breaklines]{text}
We want to create a reward function for playing the Atari game <GAME>. As a first step we want to collect functions that are helpful for understanding events that are happening in the game that could be relevant for the reward, i.e., items colliding. In the following there will be existing game objects given, please generate functions that can be used to understand the game state. Please don't use undefined variables or functions. 

Here is a description of the game and its objects:

<PARENT GAME OBJECT CLASS> 

<GAME OBJECT CLASSES> 

The game instructions are the following: 
 <INSTRUCTIONS> 
\end{minted}
\end{listing}

\begin{listing}[h]
\caption{Prompt for reward function based on identified functions from before.}
\label{lst:cot_prompt_reward}
\begin{minted}[mathescape,
               linenos,
               numbersep=5pt,
               gobble=0,
               frame=lines,
               framesep=2mm,
               fontsize=\footnotesize,
               breaksymbolleft={},
               tabsize=2,breaklines]{text}
Now please create a object centric reward function to train a reinforcement learning agent to play the game <GAME>. The reward function uses a list of objects of type GameObject as input, i.e.:
```python
def reward_function(game_objects) -> float:
    ... 
    return reward
```
You can use the identified functions from before. Please don't use other undefined variables or functions.
 <INSTRUCTIONS> 
\end{minted}
\end{listing}

\begin{listing}[h]
\caption{Prompt for rescaling reward values.}
\label{lst:reward_scaling}
\begin{minted}[mathescape,
               linenos,
               numbersep=5pt,
               gobble=0,
               frame=lines,
               framesep=2mm,
               fontsize=\footnotesize,
               breaksymbolleft={},
               tabsize=2,breaklines]{text}
"Thank you. Now please adjust the rewards so that the rewards are in the range [-1, 1]."
\end{minted}

\end{listing}

\subsubsection{The Object Properties used for OCALM}
\label{app:prop_and_feats}
In this paper, we used different object properties as the inputs to the LLM-written reward functions.
The object-centric environment context is given by the classes provided by the OCAtari framework \citep{delfosse2023ocatari}, i.e., the parent game object class\footnote{ \url{https://github.com/k4ntz/OC_Atari/blob/master/ocatari/ram/game_objects.py}} and the game-specific objects \footnote{e.g., \url{https://github.com/k4ntz/OC_Atari/blob/master/ocatari/ram/pong.py}}. The game objects related to the score were omitted (since we are assuming a reward-free environment).
An overview of the object properties used by OCALM is provided in \cref{app:concept_definitions} and concrete implementation details can be found in \cref{app:game-objects}.

\begin{table}[!ht]
\centering
\begin{tabular}{lll}
\toprule
Name               & Definition & Description \\
\midrule
class              &  \texttt{NAME}       & object class (\eg "Agent", "Ball", "Ghost")            \\
position           &  $x,y$       &   position on the screen           \\
position history   &  $x_t,y_t, x_{t-1}, y_{t-1}$       &   position and past position on the screen          \\
orientation        &  $o$       & object's orientation if available            \\
RGB                &  $R, G, B$      &  RGB values           \\
\bottomrule
\end{tabular}
\caption{Descriptions of object properties used by OCALM.}
\label{app:concept_definitions}
\end{table}

\subsubsection{Example of Game Objects}\label{app:game-objects}

As described in the previous section, the game objects are Python classes provided by the (MIT licensed) OCAtari framework \citep{delfosse2023ocatari}.
We provide the parent class for game objects and an example of game objects for Pong here for illustration purposes.
Note that we have elided parts of the parent class in the listing for brevity, indicated by \verb|#elided#|.
Refer to \url{https://github.com/k4ntz/OC_Atari/blob/v0.1.0/ocatari/ram/game_objects.py} for the full parent class
and \url{https://github.com/k4ntz/OC_Atari/blob/v0.1.0/ocatari/ram/pong.py} for the source of the Pong example.

\newenvironment{longlisting}{\captionsetup{type=listing}}{}

\begin{longlisting}
\begin{minted}[mathescape,
               linenos,
               numbersep=5pt,
               gobble=0,
               frame=lines,
               framesep=2mm,
               fontsize=\footnotesize,
               linenos,tabsize=2,breaklines]{python}
    class GameObject:
    """
    The Parent Class of every detected object in the Atari games (RAM Extraction mode)

    #elided#
    """

    GET_COLOR = False
    GET_WH = False

    def __init__(self):
        self.rgb = (0, 0, 0)
        self._xy = (0, 0)
        self.wh = (0, 0)
        self._prev_xy = None
        self._orientation = None
        self.hud = False

    def __repr__(self):
        return f"{self.__class__.__name__} at ({self._xy[0]}, {self._xy[1]}), {self.wh}"

    @property
    def category(self):
        return self.__class__.__name__

    @property
    def x(self):
        return self._xy[0]

    @property
    def y(self):
        return self._xy[1]

    #elided

    def _save_prev(self):
        self._prev_xy = self._xy

    # @x.setter
    # def x(self, x):

    #     self._xy = x, self.xy[1]
    
    # @y.setter
    # def y(self, y):
    #     self._xy = self.xy[0], y

    @property
    def orientation(self):
        return self._orientation

    @orientation.setter
    def orientation(self, o):
        self._orientation = o

    @property
    def center(self):
        return self._xy[0] + self.wh[0]/2, self._xy[1] + self.wh[1]/2

    def is_on_top(self, other):
        """
        Returns ``True`` if this and another gameobject overlap.

        :return: True if objects overlap
        :rtype: bool
        """
        return (other.x <= self.x <= other.x + other.w) and \
            (other.y <= self.y <= other.y + other.h) 
    
    def manathan_distance(self, other):
        """
        Returns the manathan distance between the center of both objects.

        :return: True if objects overlap
        :rtype: bool
        """
        c0, c1 = self.center, other.center
        return abs(c0[0] - c1[0]) + abs(c0[1]- c1[1])
    
    def closest_object(self, others):
        """
        Returns the closest object from others, based on manathan distance between the center of both objects.

        :return: (Index, Object) from others
        :rtype: int
        """
        if len(others) == 0:
            return None
        return min(enumerate(others), key=lambda item: self.manathan_distance(item[1]))


class ValueObject(GameObject):
    """
    This class represents a game object that incorporates any notion of a value.
    For example:
    * the score of the player (or sometimes Enemy).
    * the level of useable/deployable resources (oxygen bars, ammunition bars, power gauges, etc.)
    * the clock/timer

    :ivar value: The value of the score.
    :vartype value: int
    """

    def __init__(self):
        super().__init__()
        self._value = 0
        self._prev_value = None

    @property
    def value(self):
        return self._value

    @value.setter
    def value(self, value):
        self._value = None if value is None else int(value)

    @property
    def prev_value(self):
        if self._prev_value is not None:
            return self._prev_value
        else:
            return self._value

    def _save_prev(self):
        super()._save_prev()
        self._prev_value = self._value

    @property
    def value_diff(self):
        return self.value - self.prev_value
\end{minted}

\begin{minted}[mathescape,
               linenos,
               numbersep=5pt,
               gobble=0,
               frame=lines,
               framesep=2mm,
               fontsize=\footnotesize,
               linenos,tabsize=2,breaklines]{python}
class Player(GameObject):
    """
    The player figure i.e., the movable bar at the side.
    """

    def __init__(self):
        super().__init__()
        self._xy = 0, 0
        self.wh = 4, 15
        self.rgb = 92, 186, 92


class Enemy(GameObject):
    """
    The enemy bar on the opposite side.
    """

    def __init__(self):
        super().__init__()
        self._xy = 0, 0
        self.wh = 4, 15
        self.rgb = 213, 130, 74


class Ball(GameObject):
    """
    The game ball.
    """

    def __init__(self):
        super().__init__()
        self._xy = 0, 0
        self.wh = 2, 4
        self.rgb = 236, 236, 236
\end{minted}
\caption{The parent classes for game objects.}
\label{lst:game_objects_parent}
\end{longlisting}

\subsection{Game Descriptions and Rewards}\label{app:llm_results}

In the following, we provide the used game descriptions, our OCALM (full) reward functions and OCALM (no relations) reward functions for the games Pong, Freeway, Seaquest and Skiing.

\subsubsection{Freeway}\label{app:freeway-rewards}

\begin{listing}[h]
\caption{Game description of Freeway.}
\label{lst:game_desc_freeway}
\begin{minted}[mathescape,
               linenos,
               numbersep=5pt,
               gobble=0,
               frame=lines,
               framesep=2mm,
               fontsize=\footnotesize,
               breaksymbolleft={},
               tabsize=2,breaklines]{text}
You control a chicken that has to cross ten horizontal lanes of a freeway traffic.
There exist two chickens in the game, you control the left chicken, that starts at the bottom of the road and should go to the top.
Cars are traveling along the horizontal lanes and you should cross without getting run over by a car.
The screen height is 160.
\end{minted}
\end{listing}

\begin{longlisting}
\begin{minted}[mathescape,
               linenos,
               numbersep=5pt,
               gobble=0,
               frame=lines,
               framesep=2mm,
               fontsize=\footnotesize,
               linenos,tabsize=2,breaklines]{python}
from ocatari.ram.freeway import *


def detect_collision(chicken, car):
    """
    Determines if a collision occurs between a chicken and a car.

    :param chicken: An instance of the Chicken class.
    :param car: An instance of the Car class.
    :return: True if there is a collision, False otherwise.
    """
    chicken_x1, chicken_y1 = chicken.x, chicken.y
    chicken_x2, chicken_y2 = chicken.x + chicken.w, chicken.y + chicken.h
    car_x1, car_y1 = car.x, car.y
    car_x2, car_y2 = car.x + car.w, car.y + car.h

    # Check if any corner of the chicken is within the car's boundaries
    if (car_x1 <= chicken_x1 <= car_x2 or car_x1 <= chicken_x2 <= car_x2) and (
        car_y1 <= chicken_y1 <= car_y2 or car_y1 <= chicken_y2 <= car_y2
    ):
        return True
    return False


def has_reached_top(chicken, screen_height):
    """
    Checks if the chicken has reached the top of the screen.

    :param chicken: An instance of the Chicken class.
    :param screen_height: The height of the screen (int).
    :return: True if the chicken has reached the top, False otherwise.
    """
    return chicken.y <= 0


def progress_made(chicken, screen_height):
    """
    Measures the progress made by the chicken towards the top of the screen.

    :param chicken: An instance of the Chicken class.
    :param screen_height: The height of the screen (int).
    :return: A float representing the percentage of the screen crossed.
    """
    return (screen_height - chicken.y) / screen_height


def check_if_reset(chicken, initial_position):
    """
    Checks if the chicken has been reset to the initial position (usually after a collision).

    :param chicken: An instance of the Chicken class.
    :param initial_position: The starting position of the chicken (tuple of int).
    :return: True if the chicken is at the initial position, False otherwise.
    """
    return chicken.xy == initial_position


def find_closest_car(chicken, cars):
    """
    Finds the closest car to the chicken based on the Manhattan distance.

    :param chicken: An instance of the Chicken class.
    :param cars: A list of Car instances.
    :return: The closest car instance.
    """
    if not cars:
        return None
    return min(cars, key=lambda car: chicken.manathan_distance(car))


def reward_function(game_objects) -> float:
    # Initialize reward
    reward = 0.0

    # Constants
    SCREEN_HEIGHT = 160
    COLLISION_PENALTY = -1.0  # Scaled down to fit within [-1, 1]
    PROGRESS_REWARD = 0.1  # Scaled down to incrementally increase reward
    SUCCESS_REWARD = 1.0  # Maximum reward for reaching the top

    # Filter out chickens and cars from game_objects
    chickens = [obj for obj in game_objects if isinstance(obj, Chicken)]
    cars = [obj for obj in game_objects if isinstance(obj, Car)]

    # Assume control of the leftmost chicken (player's chicken)
    if chickens:
        player_chicken = min(chickens, key=lambda c: c.x)

        # Check if the chicken has reached the top
        if has_reached_top(player_chicken, SCREEN_HEIGHT):
            reward += SUCCESS_REWARD

        # Reward based on progress towards the top
        reward += progress_made(player_chicken, SCREEN_HEIGHT) * PROGRESS_REWARD

        # Check for collisions with any car
        for car in cars:
            if detect_collision(player_chicken, car):
                reward += COLLISION_PENALTY
                break  # Only penalize once per time step

    # Ensure reward stays within the range [-1, 1]
    reward = max(min(reward, 1.0), -1.0)

    return reward
\end{minted}
\caption{OCALM (full) reward function of the game Freeway.}
\label{lst:cot_freeway}
\end{longlisting}

\clearpage
\begin{longlisting}
\begin{minted}[mathescape,
               linenos,
               numbersep=5pt,
               gobble=0,
               frame=lines,
               framesep=2mm,
               fontsize=\footnotesize,
               linenos,tabsize=2,breaklines]{python}
from ocatari.ram.freeway import *

def reward_function(game_objects) -> float:
    # Initialize reward
    reward = 0.0
    
    # Define constants for screen dimensions
    SCREEN_HEIGHT = 160
    
    # Find the player's chicken and cars in the game objects
    player_chicken = None
    cars = []
    
    for obj in game_objects:
        if obj.category == "Chicken" and obj.x < SCREEN_HEIGHT // 2:  # Assuming player chicken starts on the left
            player_chicken = obj
        elif obj.category == "Car":
            cars.append(obj)
    
    # If the player's chicken is not found, return a neutral reward
    if player_chicken is None:
        return reward
    
    # Reward for moving upwards, normalized by screen height
    reward += (player_chicken.dy / SCREEN_HEIGHT)
    
    # Penalty for moving downwards, normalized and increased in impact
    if player_chicken.dy < 0:
        reward -= 2 * (abs(player_chicken.dy) / SCREEN_HEIGHT)
    
    # Check for collisions with cars
    for car in cars:
        if player_chicken.is_on_top(car):
            # Large penalty for collisions, ensuring it stays within the range
            reward -= 0.5
    
    # Bonus for reaching the top of the screen, ensuring it stays within the range
    if player_chicken.y <= 0:
        reward += 0.5
    
    # Ensure the reward is clamped between -1 and 1
    reward = max(min(reward, 1), -1)
    
    return reward
\end{minted}
\caption{Direct reward function of the game Freeway.}
\label{lst:direct_freeway}
\end{longlisting}

\clearpage

\subsubsection{Pong}\label{app:pong-rewards}

\begin{listing}[h]
\caption{Game description of Pong.}
\label{lst:game_desc_pong}
\begin{minted}[mathescape,
               linenos,
               numbersep=5pt,
               gobble=0,
               frame=lines,
               framesep=2mm,
               fontsize=\footnotesize,
               linenos,tabsize=2,breaklines]{text}
In this game the agent has to knock the ball past the enemy's paddle, situated on the left (good), and avoid letting the ball go past its paddle, on the right (bad). If the ball passes the paddle of the enemy, the agent gets a point. If the ball passes past the agents paddle, the enemy gets a point. After a scored point the ball continues moving behind the paddle for multiple frames before it respawns for a new round.
\end{minted}
\end{listing}

\begin{longlisting}
\begin{minted}[mathescape,
               linenos,
               numbersep=5pt,
               gobble=0,
               frame=lines,
               framesep=2mm,
               fontsize=\footnotesize,
               linenos,tabsize=2,breaklines]{python}
from ocatari.ram.pong import *


def check_collision(obj1, obj2):
    """
    Check if two GameObjects are colliding.

    :param obj1: First game object.
    :param obj2: Second game object.
    :return: True if objects are colliding, False otherwise.
    """
    return (
        obj1.x < obj2.x + obj2.w
        and obj1.x + obj1.w > obj2.x
        and obj1.y < obj2.y + obj2.h
        and obj1.y + obj1.h > obj2.y
    )


def ball_passed_paddle(ball, paddle, playing_field_width):
    """
    Check if the ball has passed the given paddle.

    :param ball: The ball object.
    :param paddle: The paddle object (player or enemy).
    :param playing_field_width: The width of the playing field.
    :return: True if the ball has passed the paddle, False otherwise.
    """
    if paddle.category == "Player":
        # Check if the ball has passed the player's paddle on the right
        return ball.x > playing_field_width
    elif paddle.category == "Enemy":
        # Check if the ball has passed the enemy's paddle on the left
        return ball.x + ball.w < 0
    return False


def update_game_state(objects):
    """
    Update the game state by saving the previous positions of the objects.

    :param objects: List of all game objects.
    """
    for obj in objects:
        obj._save_prev()


def detect_score_event(ball, player_paddle, enemy_paddle, playing_field_width):
    """
    Detect if a scoring event has occurred.

    :param ball: The ball object.
    :param player_paddle: The player's paddle object.
    :param enemy_paddle: The enemy's paddle object.
    :param playing_field_width: The width of the playing field.
    :return: 'player' if player scores, 'enemy' if enemy scores, None otherwise.
    """
    if ball_passed_paddle(ball, enemy_paddle, playing_field_width):
        return "player"
    elif ball_passed_paddle(ball, player_paddle, playing_field_width):
        return "enemy"
    return None


def reward_function(game_objects) -> float:
    # Initialize reward
    reward = 0.0

    # Identify the ball, player paddle, and enemy paddle from game_objects
    ball = None
    player_paddle = None
    enemy_paddle = None
    playing_field_width = 160  # Assuming a standard Atari Pong width

    for obj in game_objects:
        if isinstance(obj, Ball):
            ball = obj
        elif isinstance(obj, Player):
            player_paddle = obj
        elif isinstance(obj, Enemy):
            enemy_paddle = obj

    # Check if the ball and paddles are identified correctly
    if ball is None or player_paddle is None or enemy_paddle is None:
        return reward  # No change in reward if essential objects are missing

    # Check for scoring events
    score_event = detect_score_event(
        ball, player_paddle, enemy_paddle, playing_field_width
    )
    if score_event == "player":
        reward += 1.0  # Reward for scoring against the enemy
    elif score_event == "enemy":
        reward -= 1.0  # Penalty for the enemy scoring

    # Check for collisions with paddles to encourage hitting the ball
    if check_collision(ball, player_paddle) or check_collision(ball, enemy_paddle):
        reward += 0.1  # Small reward for hitting the ball with either paddle

    # Normalize the reward to be within [-1, 1]
    reward = max(min(reward, 1), -1)

    return reward
\end{minted}
\caption{OCALM (full)  reward function of the game Pong.}
\label{lst:cot_pong}
\end{longlisting}

\clearpage

\begin{longlisting}
\begin{minted}[mathescape,
               linenos,
               numbersep=5pt,
               gobble=0,
               frame=lines,
               framesep=2mm,
               fontsize=\footnotesize,
               linenos,tabsize=2,breaklines]{python}
from ocatari.ram.pong import *

def reward_function(game_objects) -> float:
    # Initialize reward
    reward = 0.0
    
    # Define identifiers for game objects
    player_id = None
    enemy_id = None
    ball_id = None
    
    # Find indices for player, enemy, and ball
    for i, obj in enumerate(game_objects):
        if obj.category == "Player":
            player_id = i
        elif obj.category == "Enemy":
            enemy_id = i
        elif obj.category == "Ball":
            ball_id = i
    
    # Get the game objects
    player = game_objects[player_id]
    enemy = game_objects[enemy_id]
    ball = game_objects[ball_id]
    
    # Check if the ball has passed the enemy paddle
    if ball.x < enemy.x:
        reward += 1  # Reward for scoring a point
    
    # Check if the ball has passed the player paddle
    if ball.x > player.x + player.w:
        reward -= 1  # Penalty for letting the enemy score
    
    # Normalize reward to be within the range [-1, 1]
    reward = max(min(reward, 1), -1)
    
    return reward
\end{minted}
\caption{Direct reward function of the game Pong.}
\label{lst:direct_pong}
\end{longlisting}

\clearpage

\subsubsection{Seaquest}\label{app:seaquest-rewards}

\begin{listing}[h]
\begin{minted}[mathescape,
               linenos,
               numbersep=5pt,
               gobble=0,
               frame=lines,
               framesep=2mm,
               fontsize=\footnotesize,
               linenos,tabsize=2,breaklines]{text}
You a sub (Player) able to move in all directions and fire torpedoes.
The goal is to retrieve as many divers as you can, while dodging and blasting enemy subs and killer sharks.
The game begins with one sub and three waiting on the horizon. Each time you increase your score by 10,000 points, an extra sub will be delivered to your base. 
Your sub will explode if it collides with anything except your divers.The sub has a limited amount of oxygen that decreases at a constant rate during the game. When the oxygen tank is almost empty, you need to surface and if you don't do it in time, your sub will blow up and you'll lose one diver. 
Each time you're forced to surface, with less than six divers, you lose one diver as well.
\end{minted}
\caption{Game description of Seaquest.}
\label{lst:game_desc_seaquest}
\end{listing}

\begin{longlisting}
\begin{minted}[mathescape,
               linenos,
               numbersep=5pt,
               gobble=0,
               frame=lines,
               framesep=2mm,
               fontsize=\footnotesize,
               linenos,tabsize=2,breaklines]{python}
from ocatari.ram.seaquest import *

def check_collision(obj1, obj2):
    """
    Check if two GameObjects collide based on their bounding boxes.
    """
    return (obj1.x < obj2.x + obj2.w and
            obj1.x + obj1.w > obj2.x and
            obj1.y < obj2.y + obj2.h and
            obj1.y + obj1.h > obj2.y)

def update_game_state(objects):
    """
    Update positions of all game objects and check for collisions.
    """
    collisions = []
    for obj in objects:
        # Update position based on velocity
        obj.xy = (obj.x + obj.dx, obj.y + obj.dy)
        
        # Check for collisions with other objects
        for other in objects:
            if obj != other and check_collision(obj, other):
                collisions.append((obj, other))
    return collisions

def manage_oxygen_and_lives(player, oxygen_bar, lives):
    """
    Decrease oxygen levels and manage lives based on oxygen and collisions.
    """
    # Decrease oxygen
    oxygen_bar.value -= 1
    if oxygen_bar.value <= 0:
        player.lives -= 1
        oxygen_bar.value = 100  # Reset oxygen after surfacing or losing a life
    
    # Check if lives are depleted
    if lives.value <= 0:
        print("Game Over")

def update_score_and_divers(player, divers_collected, score):
    """
    Update score based on collected divers and manage divers.
    """
    for diver in divers_collected:
        if check_collision(player, diver):
            score.value += 1000  # Increment score for each diver collected
            divers_collected.remove(diver)  # Remove diver from the game

def fire_torpedo(player, torpedoes):
    """
    Create a new torpedo at the player's location and add it to the torpedoes list.
    """
    new_torpedo = PlayerMissile()
    new_torpedo.xy = player.xy
    torpedoes.append(new_torpedo)


def reward_function(game_objects) -> float:
    reward = 0.0
    
    # Define categories for easy identification
    player = None
    divers = []
    enemies = []
    player_missiles = []
    enemy_missiles = []
    oxygen_bar = None
    
    # Classify objects
    for obj in game_objects:
        if isinstance(obj, Player):
            player = obj
        elif isinstance(obj, Diver):
            divers.append(obj)
        elif isinstance(obj, Shark) or isinstance(obj, Submarine):
            enemies.append(obj)
        elif isinstance(obj, PlayerMissile):
            player_missiles.append(obj)
        elif isinstance(obj, EnemyMissile):
            enemy_missiles.append(obj)
        elif isinstance(obj, OxygenBar):
            oxygen_bar = obj
    
    # Check for collisions and manage interactions
    if player:
        for diver in divers:
            if check_collision(player, diver):
                reward += 0.1  # Scaled down reward for collecting a diver
                divers.remove(diver)  # Assume diver is collected and removed from the game
        for enemy in enemies:
            if check_collision(player, enemy):
                reward -= 0.1  # Scaled down penalty for colliding with an enemy
        for missile in enemy_missiles:
            if check_collision(player, missile):
                reward -= 0.05  # Scaled down penalty for getting hit by an enemy missile
        # Reward for hitting enemies with missiles
        for missile in player_missiles:
            for enemy in enemies:
                if check_collision(missile, enemy):
                    reward += 0.05  # Scaled down reward for destroying an enemy
                    enemies.remove(enemy)  # Assume enemy is destroyed and removed from the game
                    player_missiles.remove(missile)  # Remove missile after hitting

    # Manage oxygen levels
    if oxygen_bar and oxygen_bar.value <= 20:
        reward -= 0.05  # Scaled down penalty for low oxygen levels
    
    # Encourage surfacing if oxygen is too low
    if oxygen_bar and oxygen_bar.value <= 10:
        reward -= 0.1  # Scaled down higher penalty for critically low oxygen

    return reward
\end{minted}
\caption{OCALM (full)  reward function of the game Seaquest.}
\label{lst:cot_seaquest}
\end{longlisting}

\clearpage

\begin{longlisting}
\begin{minted}[mathescape,
               linenos,
               numbersep=5pt,
               gobble=0,
               frame=lines,
               framesep=2mm,
               fontsize=\footnotesize,
               linenos,tabsize=2,breaklines]{python}
from ocatari.ram.seaquest import *

def reward_function(game_objects) -> float:
    # Initialize reward
    reward = 0.0

    # Define constants for reward values, adjusted to be within the range [-1, 1]
    REWARD_FOR_COLLECTING_DIVER = 0.1
    PENALTY_FOR_COLLISION = -0.1
    PENALTY_FOR_LOW_OXYGEN = -0.05
    REWARD_FOR_BLASTING_ENEMY = 0.05
    PENALTY_FOR_LOSING_DIVER_WHEN_SURFACING = -0.025

    # Helper function to find an object by its class name
    def find_objects_by_type(type_name):
        return [obj for obj in game_objects if obj.category == type_name]

    # Get specific game objects
    player = find_objects_by_type('Player')[0] if find_objects_by_type('Player') else None
    divers = find_objects_by_type('Diver')
    sharks = find_objects_by_type('Shark')
    enemy_subs = find_objects_by_type('Submarine')
    enemy_missiles = find_objects_by_type('EnemyMissile')
    player_missiles = find_objects_by_type('PlayerMissile')
    oxygen_bar = find_objects_by_type('OxygenBar')[0] if find_objects_by_type('OxygenBar') else None

    # Reward for collecting divers
    for diver in divers:
        if player and player.is_on_top(diver):
            reward += REWARD_FOR_COLLECTING_DIVER

    # Penalty for collisions with sharks or enemy submarines
    for shark in sharks:
        if player and player.is_on_top(shark):
            reward += PENALTY_FOR_COLLISION

    for enemy_sub in enemy_subs:
        if player and player.is_on_top(enemy_sub):
            reward += PENALTY_FOR_COLLISION

    # Check for low oxygen
    if oxygen_bar and oxygen_bar.value < 20:
        reward += PENALTY_FOR_LOW_OXYGEN

    # Reward for blasting enemy submarines with missiles
    for missile in player_missiles:
        for enemy_sub in enemy_subs:
            if missile.is_on_top(enemy_sub):
                reward += REWARD_FOR_BLASTING_ENEMY

    # Penalty for enemy missiles hitting the player
    for missile in enemy_missiles:
        if player and missile.is_on_top(player):
            reward += PENALTY_FOR_COLLISION

    # Penalty for surfacing with less than six divers
    collected_divers = find_objects_by_type('CollectedDiver')
    if len(collected_divers) < 6 and player and player.y == 0:  # Assuming y=0 is the surface
        reward += PENALTY_FOR_LOSING_DIVER_WHEN_SURFACING

    return reward
\end{minted}
\caption{Direct reward function of the game Seaquest.}
\label{lst:direct_seaquest}
\end{longlisting}

\clearpage

\subsubsection{Skiing}\label{app:skiing-rewards}

\begin{listing}[h]
\begin{minted}[mathescape,
               linenos,
               numbersep=5pt,
               gobble=0,
               frame=lines,
               framesep=2mm,
               fontsize=\footnotesize,
               linenos,tabsize=2,breaklines]{text}
You control a skier (Player), going down a slope who can move sideways.
The Player is at the top of the screen, staying at the same y position but the other objects of the environments are moving up towards him. 
The goal is to ski in between the horizontal pairs of flags. 
There can be up to two pairs of poles on the screen.
Do not hit a tree or a flag or you'll fall and lose time.
\end{minted}
\caption{Game description of Skiing.}
\label{lst:game_desc_skiing}
\end{listing}

\begin{longlisting}
\begin{minted}[mathescape,
               linenos,
               numbersep=5pt,
               gobble=0,
               frame=lines,
               framesep=2mm,
               fontsize=\footnotesize,
               linenos,tabsize=2,breaklines]{python}
from ocatari.ram.skiing import *

def check_collision(player, objects):
    """
    Check if the player has collided with any of the given objects (flags or trees).

    :param player: The player object.
    :param objects: A list of game objects (flags or trees).
    :return: True if a collision is detected, False otherwise.
    """
    for obj in objects:
        if (player.x < obj.x + obj.w and
            player.x + player.w > obj.x and
            player.y < obj.y + obj.h and
            player.y + player.h > obj.y):
            return True
    return False

def check_gate_passage(player, flag1, flag2):
    """
    Check if the player has passed between two flags.

    :param player: The player object.
    :param flag1: The first flag object.
    :param flag2: The second flag object.
    :return: True if the player has passed between the flags, False otherwise.
    """
    if flag1.y == flag2.y:  # Ensure flags are aligned horizontally
        gate_left = min(flag1.x, flag2.x)
        gate_right = max(flag1.x + flag1.w, flag2.x + flag2.w)
        player_center_x = player.x + player.w / 2
        return gate_left <= player_center_x <= gate_right
    return False

def distance_to_nearest_obstacle(player, obstacles):
    """
    Calculate the Manhattan distance from the player to the nearest obstacle.

    :param player: The player object.
    :param obstacles: A list of obstacle objects (trees or moguls).
    :return: The Manhattan distance to the closest obstacle.
    """
    closest_distance = float('inf')
    for obstacle in obstacles:
        distance = player.manathan_distance(obstacle)
        if distance < closest_distance:
            closest_distance = distance
    return closest_distance

def update_game_state(objects, dy):
    """
    Update the positions of all game objects to simulate the player moving down the slope.

    :param objects: A list of all movable game objects.
    :param dy: The change in y-coordinate to apply to each object.
    """
    for obj in objects:
        if not obj.hud:
            obj.y += dy


def reward_function(objects):
    """
    Calculate the reward for the current game state based on the player's interactions with game objects,
    with rewards scaled to be within the range [-1, 1].

    :param objects: A list of all game objects currently in the game.
    :return: A numerical reward value.
    """
    # Initialize reward
    reward = 0

    # Define penalty and reward values, scaled to ensure the total reward remains within [-1, 1]
    collision_penalty = -1
    gate_pass_reward = 0.5
    close_to_obstacle_penalty = -0.01

    # Filter objects by type
    player = next(obj for obj in objects if isinstance(obj, Player))
    flags = [obj for obj in objects if isinstance(obj, Flag)]
    trees = [obj for obj in objects if isinstance(obj, Tree)]
    moguls = [obj for obj in objects if isinstance(obj, Mogul)]

    # Check for collisions with trees or flags
    if check_collision(player, trees + flags):
        reward += collision_penalty

    # Check for successful gate passage
    # Assuming flags come in pairs, sorted by y and then x
    flags_sorted = sorted(flags, key=lambda f: (f.y, f.x))
    for i in range(0, len(flags_sorted), 2):
        if i+1 < len(flags_sorted) and check_gate_passage(player, flags_sorted[i], flags_sorted[i+1]):
            reward += gate_pass_reward

    # Penalize being too close to the nearest obstacle
    nearest_obstacle_distance = distance_to_nearest_obstacle(player, trees + moguls)
    # Assuming a threshold below which the player is considered too close to an obstacle
    if nearest_obstacle_distance < 20:
        reward += close_to_obstacle_penalty * (20 - nearest_obstacle_distance)

    # Ensure the reward is within the range [-1, 1]
    reward = max(min(reward, 1), -1)

    return reward
\end{minted}
\caption{OCALM (full)  reward function of the game Skiing.}
\label{lst:cot_skiing}
\end{longlisting}

\clearpage

\begin{longlisting}
\begin{minted}[mathescape,
               linenos,
               numbersep=5pt,
               gobble=0,
               frame=lines,
               framesep=2mm,
               fontsize=\footnotesize,
               linenos,tabsize=2,breaklines]{python}
from ocatari.ram.skiing import *

def reward_function(game_objects) -> float:
    # Initialize reward
    reward = 0.0
    
    # Define constants for reward/penalty values
    FLAG_PASS_REWARD = 0.1
    TREE_COLLISION_PENALTY = -0.3
    FLAG_COLLISION_PENALTY = -0.2
    MOGUL_COLLISION_PENALTY = -0.05
    
    # Helper function to check if the player collides with any object
    def check_collision(player, obj):
        return (obj.x <= player.x <= obj.x + obj.w or obj.x <= player.x + player.w <= obj.x + obj.w) and \
               (obj.y <= player.y <= obj.y + obj.h or obj.y <= player.y + player.h <= obj.y + obj.h)
    
    # Extract player and other objects
    player = None
    flags = []
    trees = []
    moguls = []
    
    for obj in game_objects:
        if isinstance(obj, Player):
            player = obj
        elif isinstance(obj, Flag):
            flags.append(obj)
        elif isinstance(obj, Tree):
            trees.append(obj)
        elif isinstance(obj, Mogul):
            moguls.append(obj)
    
    # Check for collisions and successful flag passes
    if player:
        # Check for collisions with trees
        for tree in trees:
            if check_collision(player, tree):
                reward += TREE_COLLISION_PENALTY
        # Check for collisions with flags
        for flag in flags:
            if check_collision(player, flag):
                reward += FLAG_COLLISION_PENALTY
        # Check for collisions with moguls
        for mogul in moguls:
            if check_collision(player, mogul):
                reward += MOGUL_COLLISION_PENALTY
        # Check if player passes between flags (assuming flags come in pairs)
        if len(flags) >= 2:
            # Sort flags by x to pair them
            sorted_flags = sorted(flags, key=lambda f: f.x)
            for i in range(0, len(sorted_flags) - 1, 2):
                flag1 = sorted_flags[i]
                flag2 = sorted_flags[i+1]
                # Check if player is between the flags
                if flag1.x < player.x < flag2.x:
                    reward += FLAG_PASS_REWARD
    
    return reward
\end{minted}
\caption{Direct reward function of the game Skiing.}
\label{lst:direct_skiing}
\end{longlisting}

\end{document}